\documentclass[journal]{IEEEtran}
\usepackage{hyperref}
\usepackage{url}
\usepackage{amsthm}
\usepackage{amsmath}
\usepackage{amssymb}
\usepackage{graphicx}
\usepackage{color}
\usepackage{xfrac}
\usepackage[export]{adjustbox}
\usepackage{verbatim}
\usepackage[normalem]{ulem}

\definecolor{darkblue}{rgb}{0,0,0.5}
\definecolor{darkgreen}{rgb}{0,0.66,0.0}
\definecolor{gray}{rgb}{0.33,0.33,0.33}
\definecolor{somegreenalex}{rgb}{0.0,0.8,0.0}

\newcommand{\x}{\boldsymbol{x}}
\newcommand{\h}{\boldsymbol{h}}
\renewcommand{\r}{\boldsymbol{r}}

\begin{document}
\title{Evaluating the visualization of what a\\ Deep Neural Network has learned}

\author{Wojciech Samek$^\dag$~\IEEEmembership{Member,~IEEE,}
        Alexander Binder$^\dag$,
	Gr\'egoire Montavon,
        Sebastian Bach, 
        and~Klaus-Robert M\"{u}ller,~\IEEEmembership{Member,~IEEE,}
\thanks{This work was supported by the Brain Korea 21 Plus Program through the National Research Foundation of Korea funded by the Ministry of Education. This work was also supported by the grant DFG (MU~987/17-1) and by the German Ministry for Education and Research as Berlin Big Data Center BBDC (01IS14013A). This publication only reflects the authors views. Funding agencies are not liable for any use that may be made of the information contained herein. {\it Asterisks indicate corresponding author}.}
\thanks{$^*$W. Samek is with Fraunhofer Heinrich Hertz Institute, 10587 Berlin, Germany. (e-mail: wojciech.samek@hhi.fraunhofer.de)}
\thanks{$^*$A. Binder is with the ISTD Pillar, Singapore University of Technology and Design (SUTD), Singapore, and with the Berlin Institute of Technology (TU Berlin), 10587 Berlin, Germany. (e-mail: alexander\_binder@sutd.edu.sg)}
\thanks{G. Montavon is with the Berlin Institute of Technology (TU Berlin), 10587 Berlin, Germany. (e-mail: gregoire.montavon@tu-berlin.de)}
\thanks{S. Bach is with Fraunhofer Heinrich Hertz Institute, 10587 Berlin, Germany. (e-mail: sebastian.bach@hhi.fraunhofer.de)}
\thanks{$^*$K.-R. M\"uller is with the Berlin Institute of Technology (TU Berlin), 10587 Berlin, Germany, and also with the Department of Brain and Cognitive Engineering, Korea University, Seoul 136-713, Korea (e-mail: klaus-robert.mueller@tu-berlin.de)}
\thanks{$^\dag$ WS and AB contributed equally}}
\markboth{Samek et al. $-$ Evaluating the visualization of what a Deep Neural Network has learned}{Samek et al. $-$ Evaluating the visualization of what a Deep Neural Network has learned}

\maketitle

\begin{abstract}
Deep Neural Networks (DNNs) have demonstrated impressive performance in complex machine learning tasks such as image classification or speech recognition. However, due to their multi-layer nonlinear structure, they are not transparent, i.e., it is hard to grasp {\it what} makes them arrive at a particular classification or recognition decision given a new unseen data sample. Recently, several approaches have been proposed enabling one to understand and interpret the reasoning embodied in a DNN for a single test image. These methods quantify the ``importance'' of individual pixels wrt the classification decision and allow a visualization in terms of a heatmap in pixel/input space. While the usefulness of heatmaps can be judged subjectively by a human, an objective quality measure is missing. In this paper we present a general methodology based on region perturbation for evaluating ordered collections of pixels such as heatmaps. 
We compare heatmaps computed by three different methods on the SUN397, ILSVRC2012 and MIT Places data sets. Our main result is that the recently proposed Layer-wise Relevance Propagation (LRP) algorithm qualitatively and quantitatively provides a better explanation of what made a DNN arrive at a particular classification decision than the sensitivity-based approach or the deconvolution method. We provide theoretical arguments to explain this result and discuss its practical implications.
Finally, we investigate the use of heatmaps for unsupervised assessment of neural network performance.
\end{abstract}


\section{Introduction}
Deep Neural Networks (DNNs) are powerful methods for solving large scale real world problems such as automated image classification \cite{DBLP:conf/nips/KrizhevskySH12, DBLP:conf/nips/CiresanGGS12, DBLP:journals/corr/SzegedyLJSRAEVR14, ciresan2012multi}, natural language processing \cite{DBLP:journals/jmlr/CollobertWBKKK11,Socher-etal:2013}, human action recognition \cite{DBLP:conf/icml/JiXYY10,DBLP:conf/cvpr/LeZYN11}, or physics \cite{montavon-njp13}; see also \cite{MonMue12b}.
Since DNN training methodologies (unsupervised pretraining, dropout, parallelization, GPUs etc.) have been improved \cite{lecun2012efficient}, DNNs are recently able to harvest extremely large amounts of training data and can thus achieve record performances in many research fields. At the same time, DNNs are generally conceived as black box methods, and users might consider this lack of transparency a drawback in practice. Namely, it is difficult to intuitively and quantitatively understand the result of DNN inference, i.e. for an {\em individual} novel input data point, {\it what} made the trained DNN model arrive at a particular response. Note that this aspect differs from feature selection \cite{guyon2003introduction}, where the question is: which features are on average salient for the {\em ensemble} of training data.

Only recently, the transparency problem has been receiving more attention for general nonlinear estimators \cite{BraBuhMue08,DBLP:journals/jmlr/BaehrensSHKHM10, hansen2011visual}. Several methods have been developed to understand what a DNN has learned \cite{erhan2010understanding, MonBraMue11, MonBraKruMue13}. While in DNN a large body of work is dedicated to visualize particular neurons or neuron layers \cite{DBLP:conf/nips/KrizhevskySH12,DBLP:conf/eccv/ZeilerF14,DBLP:journals/corr/MahendranV14,DBLP:journals/corr/SzegedyZSBEGF13,DBLP:journals/corr/GoodfellowSS14,DBLP:journals/corr/YosinskiCNFL15,DBLP:journals/corr/DosovitskiyB15}, we focus here on methods which visualize the impact of particular regions of a given and fixed single image for a prediction of this image. Zeiler and Fergus \cite{DBLP:conf/eccv/ZeilerF14} have proposed in their work a network propagation technique to identify patterns in a given input image that are linked to a particular DNN prediction. This method runs a backward algorithm that reuses the weights at each layer to propagate the prediction from the output down to the input layer, leading to the creation of meaningful patterns in input space. This approach was designed for a particular type of neural network, namely convolutional nets with max-pooling and rectified linear units. A limitation of the deconvolution method is the absence of a particular theoretical criterion that would directly connect the predicted output to the produced pattern in a quantifiable way. Furthermore, the usage of image-specific information for generating the backprojections in this method is limited to max-pooling layers alone. Further previous work has focused on understanding non-linear learning methods such as DNNs or kernel methods \cite{DBLP:journals/jmlr/BaehrensSHKHM10, DBLP:conf/biostec/RasmussenSMLYSH12,DBLP:journals/corr/SimonyanVZ13} essentially by sensitivity analysis in the sense of scores based on partial derivatives at the given sample. Partial derivatives look at local sensitivities detached from the decision boundary of the classifier. Simonyan et al.\ \cite{DBLP:journals/corr/SimonyanVZ13} applied partial derivatives for visualizing input sensitivities in images classified by a deep neural network. Note that although \cite{DBLP:journals/corr/SimonyanVZ13} describes a Taylor series, it relies on partial derivatives at the given image for computation of results. In a strict sense partial derivatives do not explain a classifier's decision ({\it ``what speaks for the presence of a car in the image''}), but rather tell us {\it what change would make the image more or less belong to the category car}. As shown later these two types of explanations lead to very different results in practice. An approach, Layer-wise Relevance Propagation (LRP), which is applicable to arbitrary types of neural unit activities (even if they are non-continuous) and to general DNN architectures has been proposed by Bach et al.~\cite{bach15}. This work aims at explaining the difference of a prediction $f(x)$ relative to the neutral state $f(x)=0$. The LRP method relies on a conservation principle to propagate the prediction back without using gradients. This principle ensures that the network output activity is fully redistributed through the layers of a DNN onto the input variables, i.e., neither positive nor negative evidence is lost. 

\begin{figure}[t]
\centering
\includegraphics[width=1\columnwidth]{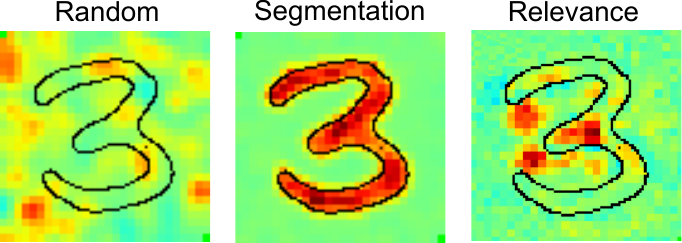}
\vskip -2mm
\caption{Comparison of three exemplary heatmaps for the image of a `3'. \underline{Left}: The randomly generated heatmap lacks interpretable information. \underline{Middle}: The segmentation heatmap focuses on the whole digit without indicating what parts of the image were particularly relevant for classification. Since it does not suffice to consider only the highlighted pixels for distinguishing an image of a `3' from images of an `8' or a `9', this heatmap is not useful for explaining classification decisions. \underline{Right}: A relevance heatmap indicates which parts of the image are used by the classifier. Here the heatmap reflects human intuition very well because the horizontal bar together with the missing stroke on the left are strong evidence that the image depicts a `3' and not any other digit.}
\label{fig:fig1}
\end{figure}

In the following we will denote the visualizations produced by the above methods as heatmaps.
While per se a heatmap is an interesting and intuitive tool that can already allow to achieve transparency, it is difficult to quantitatively evaluate the quality of a heatmap.  In other words we may ask: what exactly makes a ``good'' heatmap. A human may be able to intuitively assess the quality of a heatmap, e.g., by matching with a prior of what is regarded as being relevant (see Figure \ref{fig:fig1}). For practical applications, however, an automated objective and quantitative measure for assessing heatmap quality becomes necessary. Note that the validation of heatmap quality is important if we want to use it as input for further analysis. For example we could run computationally more expensive algorithms only on relevant regions in the image, where relevance is detected by a heatmap. 

\noindent In this paper we contribute by
\begin{itemize}
\item pointing to the issue of how to objectively evaluate the quality of heatmaps. To the best of our knowledge this question has not been raised so far.
\item introducing a generic framework for evaluating heatmaps which extends the approach in \cite{bach15} from binary inputs to color images.
\item comparing three different heatmap computation methods on three large data-sets and noting that the relevance-based LRP algorithm \cite{bach15} is more suitable for explaining the classification decisions of DNNs than the sensitivity-based approach \cite{DBLP:journals/corr/SimonyanVZ13} and the deconvolution method \cite{DBLP:conf/eccv/ZeilerF14}.
\item investigating the use of heatmaps for assessment of neural network performance.
\end{itemize}

The next section briefly introduces three existing methods for computing heatmaps. Section \ref{sec:heatmap} discusses the heatmap evaluation problem and presents a generic framework for this task. Two experimental results are presented in Section \ref{sec:results}: The first experiment compares different heatmapping algorithms on SUN397 \cite{DBLP:conf/cvpr/XiaoHEOT10}, ILSVRC2012 \cite{ILSVRC15} and MIT Places \cite{zhou2014learning} data sets and the second experiment investigates the correlation between heatmap quality and neural network performance on the CIFAR-10 data set \cite{krizhevsky2009learning}. We conclude the paper in Section \ref{sec:conclusion} and give an outlook.

\section{Understanding DNN Prediction}
\label{sec:lrp}
In the following we focus on images, but the presented techniques are applicable to any type of input domain whose elements can be processed by a neural network.

Let us consider an image $\x \in \mathbb{R}^d$, decomposable as a set of pixel values $\x = \{x_p\}$ where $p$ denotes a particular pixel, and a classification function $f: \mathbb{R}^d \to \mathbb{R}^+$. The function value $f(\x)$ can be interpreted as a score indicating the certainty of the presence of a certain type of object(s) in the image. Such functions can be learned very well by a deep neural network. Throughout the paper we assume neural networks to consist of multiple layers of neurons, where neurons are activated as
\begin{align}
& a_j^{(l+1)} = \sigma\Big(\sum_i z_{ij} + b_j^{(l+1)}\Big)\\[+3px]
& \mathrm{with} \quad z_{ij} = a_i^{(l)} w_{ij}^{(l,l+1)}
\label{eq:active}
\end{align} 
The sum operator runs over all lower-layer neurons that are connected to neuron $j$, where $a_i^{(l)}$ is the activation of a neuron $i$ in the previous layer, and where $z_{ij}$ is the contribution of neuron $i$ at layer $l$ to the activation of the neuron $j$ at layer $l+1$. The function $\sigma$ is a nonlinear monotonously increasing activation function, $w_{ij}^{(l,l+1)}$ is the weight and $b_j^{(l+1)}$ is the bias term.

\begin{figure*}[!ht]
\centering
\includegraphics[width=1.0\textwidth]{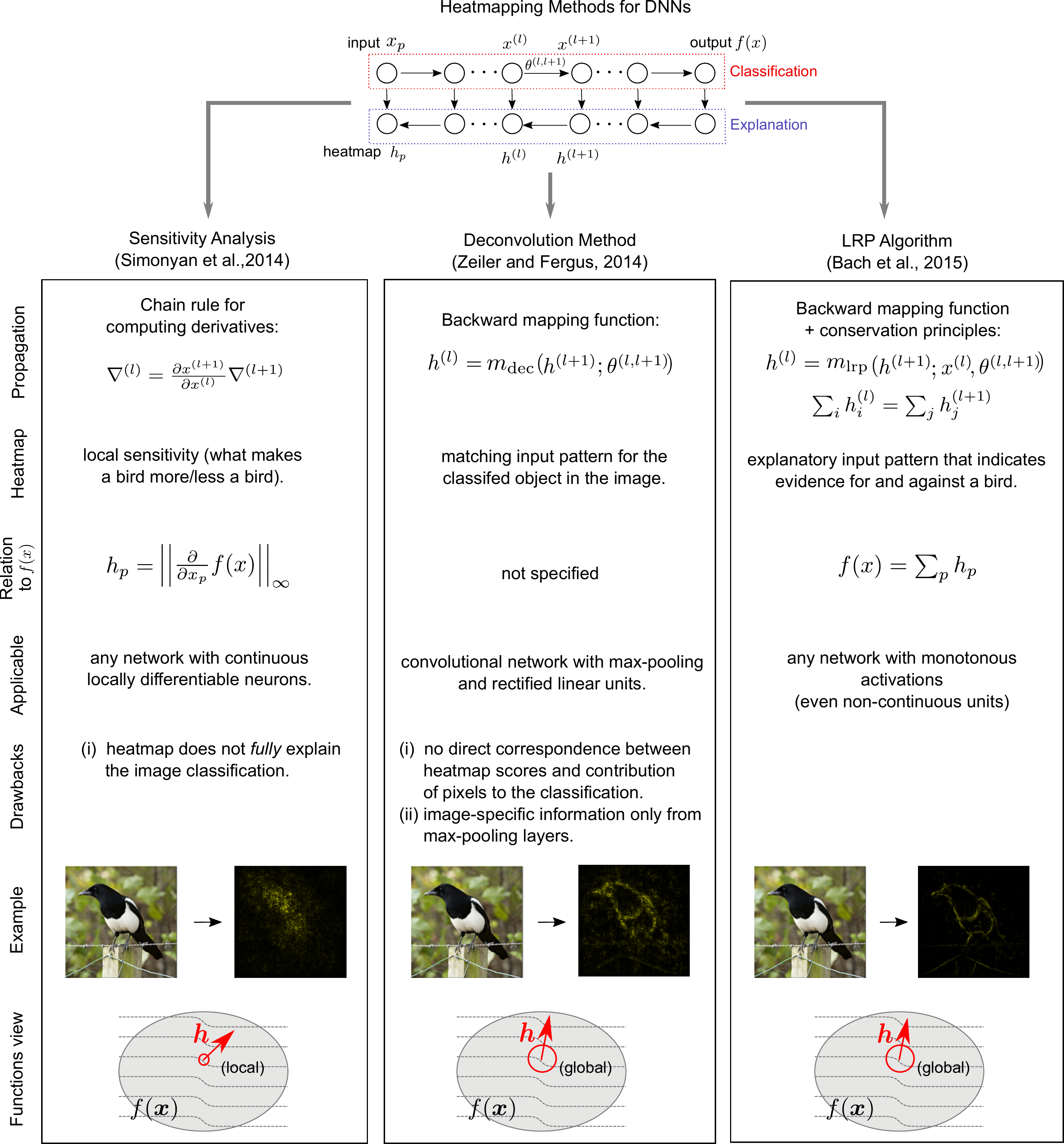}
\vskip -2mm
\caption{Comparison of the three heatmap computation methods used in this paper. \underline{Left}: Sensitivity heatmaps are based on partial derivatives, i.e., measure which pixels, when changed, would make the image belong less or more to a category (local explanations). The method is applicable to generic architectures with differentiable units. \underline{Middle}: The deconvolution method applies a convolutional network $g$ to the output of another convolutional network $f$. Network $g$ is constructed in a way to ``undo'' the operations performed by $f$. Since negative evidence is discarded and scores are not normalized during the backpropagation, the relation between heatmap scores and the classification output $f(x)$ is unclear. \underline{Right}: Layer-wise Relevance Propagation (LRP) exactly decomposes the classification output $f(x)$ into pixel relevances by observing the layer-wise conservation principle, i.e., evidence for or against a category is not lost. The algorithm does not use gradients and is therefore applicable to generic architectures (including nets with non-continuous units). LRP globally explains the classification decision and heatmap scores have a clear interpretation as evidence for or against a category.}
\label{fig:fig2}
\end{figure*}

 A \emph{heatmap} $\h = \{h_p\}$ assigns each pixel $p$ a value  $h_p = \mathcal{H}(\x,f,p)$ according to some function $\mathcal{H}$, typically derived from a class discriminant $f$. Since $\h$ has the same dimensionality as $\x$, it can be visualized as an image.
In the following we review three recent methods for computing heatmaps, all of them performing a backward propagation pass on the network: (1) a sensitivity analysis based on neural network partial derivatives, (2) the so-called deconvolution method and (3) the layer-wise relevance propagation algorithm. Figure \ref{fig:fig2} briefly summarizes the methods.

\subsection{Sensitivity Heatmaps}
\label{section:sensitivity}
A well-known tool for interpreting non-linear classifiers is sensitivity analysis \cite{DBLP:journals/jmlr/BaehrensSHKHM10}. It was used by Simonyan et al.~\cite{DBLP:journals/corr/SimonyanVZ13} to compute saliency maps of images classified by neural networks. In this approach the sensitivity of a pixel $h_p$ is computed by using the norm $\|\cdot\|_{\ell_q}$ over partial derivatives (\cite{DBLP:journals/corr/SimonyanVZ13} used $q=\infty$) for the color channel $c$ of a pixel $p$:
\begin{align}
h_p = \left\| \left( \frac{\partial}{\partial x_{p,c}}  f(\x) \right)_{c \in (r,g,b)} \right\|_{\ell_q}
\label{eq:sensitivity}
\end{align}
This quantity measures how much small changes in the pixel value locally affect the network output. Large values of $h_p$ denote pixels which largely affect the classification function $f$ if changed. Note that the direction of change (i.e., sign of the partial derivative) is lost when using the norm. 
Partial derivatives are obtained efficiently by running the backpropagation algorithm \cite{rumelhart86} throughout the multiple layers of the network. The backpropagation rule from one layer to another layer, where $x^{(l)}$ and $x^{(l+1)}$ denote the neuron activities at two consecutive layers is given by:
\begin{align}
\frac{\partial f}{\partial x^{(l)}} = \frac{\partial x^{(l+1)}}{\partial x^{(l)}} \frac{\partial f}{\partial x^{(l+1)}}
\end{align}
The backpropagation algorithm performs the following operations in the various layers:\\[+3px]
\textbf{Unpooling}: The gradient signal is redirected onto the input neuron(s) to which the corresponding output neuron is sensitive. In the case of max-pooling, the input neuron in question is the one with maximum activation value.\\[+3px]
\textbf{Nonlinearity}: Denoting $z^{(l)}_i$ is the preactivation of the $i$th neuron of the $l$th layer, backpropagating the signal through a rectified linear unit (ReLU) defined by the map $z^{(l)}_i \to \max(0,z^{(l)}_i)$ corresponds to multiplying the backpropagated gradient signal by the indicator function $1_{\{z^{(l)}_i > 0\}}$.\\[+3px]
\textbf{Filtering}: The gradient signal is convolved by a transposed version of the convolutional filter used in the forward pass.\\[+3px]
Of particular interest, the multiplication of the signal by an indicator function in the rectification layer makes the backward mapping discontinuous, and consequently strongly local. Thus, the gradient on which the heatmap is based is expected to be mostly composed of local features (e.g. what makes a given car look more/less like a car) and few global features (e.g. what are all the features that compose a given car). Note that the gradient gives for every pixel a direction in RGB-space in which the prediction increases or decreases, but it does not indicate directly to whether a particular region contains evidence for or against the prediction made by a classifier. We compute heatmaps by using Eq.~\ref{eq:sensitivity} with the norms $q = \{2, \infty\}$.

\subsection{Deconvolution Heatmaps}
\label{section:deconvolution}
Another method for heatmap computation was proposed in \cite{DBLP:conf/eccv/ZeilerF14} and uses a process termed deconvolution.  Similarly to the backpropagation method to compute the function's gradient, the idea of the deconvolution approach is to map the activations from the network's output back to pixel space using a backpropagation rule
$$
R^{(l)} = m_{\mathrm{dec}}(R^{(l+1)};\theta^{(l,l+1)}).
$$
Here, $R^{(l)},R^{(l+1)}$ denote the backward signal as it is backpropagated from one layer to the previous layer, $m_{\mathrm{dec}}$ is a predefined function that may be different for each layer and $\theta^{(l,l+1)}$ is the set of parameters connecting two layers of neurons. This method was designed for a convolutional net with max-pooling and rectified linear units, but it could also be adapted in principle for other types of architectures. The following set of rules is applied to compute deconvolution heatmaps.\\[+3px]
\textbf{Unpooling}: The locations of the maxima within each pooling region are recorded and these recordings are used to place the relevance signal from the layer above into the appropriate locations. For deconvolution this seems to be the only place besides the classifier output where image information from the forward pass is used, in order to arrive at an image-specific explanation.\\[+3px]
\textbf{Nonlinearity}: The relevance signal at a ReLU layer is passed through a ReLU function during the deconvolution process.\\[+3px]
\textbf{Filtering}: In a convolution layer, the transposed versions of the trained filters are used to backpropagate the relevance signal. This projection does not depend on the neuron activations $x^{(l)}$. \\[+3px]
The unpooling and filtering rules are the same as those derived from gradient propagation (i.e. those used in Section \ref{section:sensitivity}). The propagation rule for the ReLU nonlinearity differs from backpropagation: Here, the backpropagated signal is not multiplied by a discontinuous indicator function, but is instead passed through a rectification function similar to the one used in the forward pass. Note that unlike the indicator function, the rectification function is continuous. This continuity in the backward mapping procedure enables the capture of more global features that can be in principle useful to represent evidence for the whole object to be predicted. Note also that the deconvolution method only implicitly takes into account properties of individual images through the unpooling operation. The backprojection over filtering layers is independent of the individual image. Thus, when applied to neural networks without a pooling layer, the deconvolution method will not provide individual (image specific) explanations, but rather average  salient features (see Figure \ref{fig:fail}). Note also that negative evidence ($R^{(l+1)} < 0$) is discarded during the backpropagation due to the application of the ReLU function. Furthermore the backward signal is not normalized layer-wise, so that few dominant $R^{(l)}$ may largely determine the final heatmap scores. Due to the suppression of negative evidence and the lack of normalization the relation between the heatmap scores and the classification output cannot be expressed analytically but is implicit to the above algorithmic procedure.

For deconvolution we apply the same color channel pooling methods (2-norm, $\infty$-norm) as for sensitivity analysis. 

\subsection{Relevance Heatmaps}
\label{section:lrp}
Layer-wise Relevance Propagation (LRP) \cite{bach15} is a principled approach to decompose a classification decision into pixel-wise {\it relevances} indicating the contributions of a pixel to the overall classification score. The approach is derived from a layer-wise conservation principle \cite{bach15}, which forces the propagated quantity (e.g. evidence for a predicted class) to be preserved between neurons of two adjacent layers. Denoting by $R_{i}^{(l)}$ the relevance associated to the $i$th neuron of layer $l$ and by $R_{j}^{(l+1)}$ the relevance associated to the $j$th neuron in the next layer, the conservation principle requires that
\begin{align}
\textstyle \sum_{i} R_{i}^{(l)} = \textstyle \sum_j R_{j}^{(l+1)}
\label{eq:conservation}
\end{align}
where the sums run over all neurons of the respective layers. Applying this rule repeatedly for all layers, the heatmap resulting from LRP satisfies $\sum_p h_p = f(\x)$ where $h_p  = R_{p}^{(1)}$ and is said to be consistent with the evidence for the predicted class. Stricter definitions of conservation that involve only subsets of neurons can further impose that relevance is locally redistributed in the lower layers. The propagation rules for each type of layers are given below:\\[+3px]
\textbf{Unpooling}: Like for the previous approaches, the backward signal is redirected proportionally onto the location for which the activation was recorded in the forward pass.\\[+3px]
\textbf{Nonlinearity}: The backward signal is simply propagated onto the lower layer, ignoring the rectification operation. Note that this propagation rule satisfies Equation \ref{eq:conservation}.\\[+3px]
\textbf{Filtering}: Bach et al.\ \cite{bach15} proposed two relevance propagation rules for this layer, that satisfy Equation \ref{eq:conservation}. Let $z_{ij} = a_i^{(l)} w_{ij}^{(l,l+1)}$ be the weighted activation of neuron $i$ onto neuron $j$ in the next layer. The first rule is given by:
\begin{align}
R_{i}^{(l)} = \sum_{j} \frac{z_{ij}}{\sum_{i'} z_{i'j}+\epsilon \,\mathrm{sign}(\sum_{i'} z_{i'j}) } R_j^{(l+1)} \label{eq:lrp-basic}
\end{align}
The intuition behind this rule is that lower-layer neurons that mostly contribute to the activation of the higher-layer neuron receive a larger share of the relevance $R_j$ of the neuron $j$. The neuron $i$ then collects the relevance associated to its contribution from all upper-layer neurons $j$. A downside of this propagation rule (at least if $\epsilon = 0$) is that the denominator may tend to zero if lower-level contributions to neuron $j$ cancel each other out. The numerical instability can be overcome by either setting $\epsilon > 0$. However in that case, the conservation idea is relaxated in order to gain better numerical properties. A way to achieve exact conservation is by separating the positive and negative activations in the relevance propagation formula, which yield the second formula:
\begin{align}
R_{i}^{(l)} = \sum_{j} \left( \alpha \cdot \frac{z_{ij}^+}{\sum_{i'} z_{i'j}^+} + \beta \cdot \frac{z_{ij}^-}{\sum_{i'}z_{i'j}^-} \right) R^{(l+1)}_j \label{eq:lrp-alphabeta}.
\end{align}
Here, $z_{ij}^+$ and $z_{ij}^-$ denote the positive and negative part of $z_{ij}$ respectively, such that $z_{ij}^+ + z_{ij}^- = z_{ij}$. We enforce $\alpha + \beta = 1$ in order for the relevance propagation equations to be conservative layer-wise. It should be emphasized that unlike gradient-based techniques, the LRP formula is applicable to non-differentiable neuron activation functions. In the experiments section we use for consistency the same settings as in \cite{bach15} without having optimized the parameters, namely the LRP variant from Equation \eqref{eq:lrp-alphabeta} with $\alpha = 2$ and $\beta = -1$ (which will be denoted as LRP in subsequent figures), and twice LRP from Equation \eqref{eq:lrp-basic} with $\epsilon=0.01$ and $\epsilon=100$.\\[+3px]

Same as for the deconvolution heatmap, the LRP algorithm does not multiply its backward signal by a discontinuous function. Therefore, relevance heatmaps also favor the emergence of global features, that allow for full explanation of the class to be predicted.
In addition, a heatmap produced by LRP has the following technical advantages over sensitivity and deconvolution: (1) Localized relevance conservation ensures a proper global redistribution of relevance in the pixel space. (2) By summing relevance on each color channel, heatmaps can be directly interpreted as a measure of total relevance per pixel, without having to compute a norm. This allows for \emph{negative evidence} (i.e. parts of the image that speak against the neural network classification decision). (3) Finally LRP's rule for filtering layers takes into account \emph{both} filter weights and lower-layer neuron activations. This allows for individual explanations even in a neural network without pooling layers.

\begin{figure*}[t]
\centering
\includegraphics[width=0.8\linewidth]{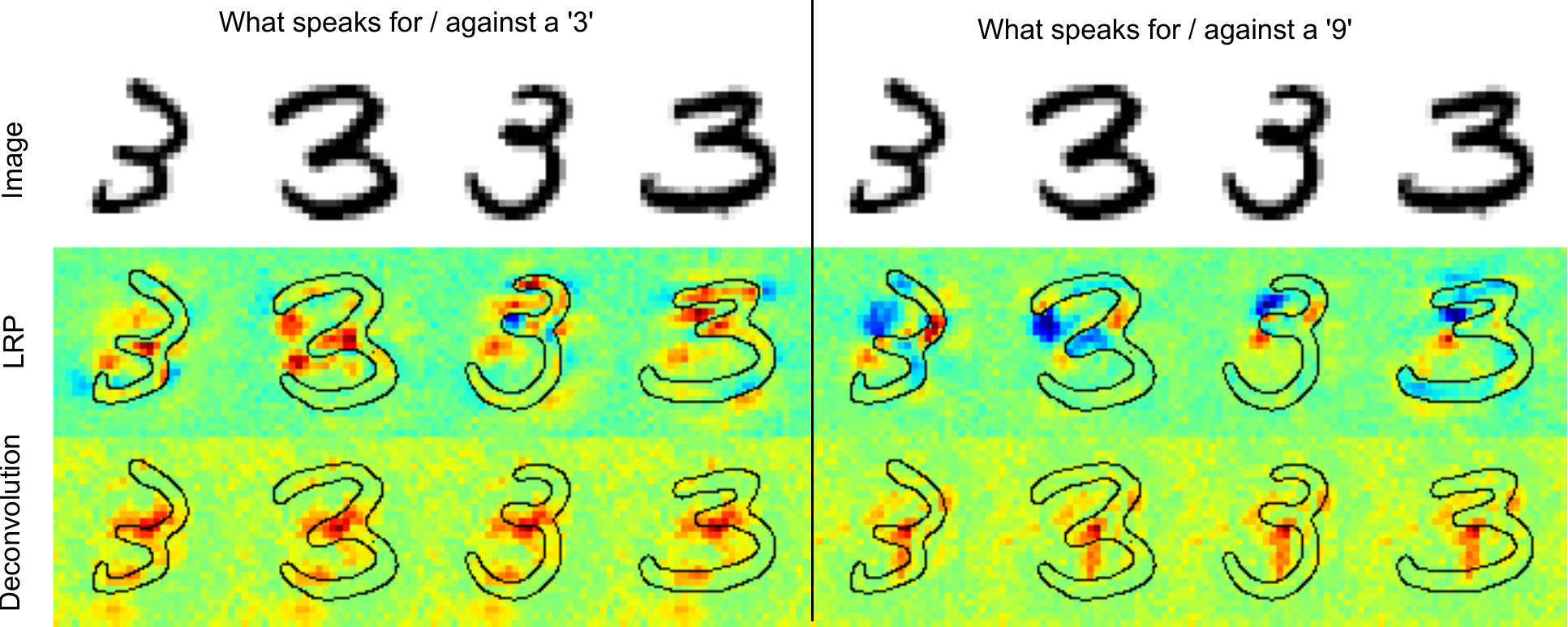}
\vskip -2mm
\caption{Comparison of LRP and deconvolution heatmaps for a neural network without pooling layers. The heatmaps in the left panel explain why the image was classified as `3', whereas heatmaps on the right side explain the classification decision `9'. The LRP heatmaps (second row) visualize both positive and negative evidence and are tailored to each individual image. The deconvolution method does not use neural activations, thus fails to provide image specific heatmaps for this network (last row). The heatmaps also do not allow to distinguish between positive and negative evidence. Since negative evidence is discarded when backpropagating through the ReLU layer, deconvolution heatmaps are not able to explain what speaks {\it against} the classification decision `9'.}
\label{fig:fail}
\end{figure*}

To demonstrate these advantages on a simple example, we compare the explanations provided by the deconvolution method and LRP for a neural network without pooling layers trained on the MNIST data set (see \cite{bach15} for details). One can see in Figure \ref{fig:fail} that LRP provides individual explanations for all images in the sense that when the digit in the image is slightly rotated, then the heatmap adapts to this rotation and highlights the relevant regions of this particular rotated digit. 
The deconvolution heatmap on the other hand is not image-specific because it only depends on the weights and not the neuron activations. If pooling layers were present in the network, then the deconvolution approach would implicitly adapt to the specific image through the unpooling operation. Still we consider this information important to be included when backprojecting over filtering layers, because neurons with large activations for a specific image should be regarded as more relevant, thus should backproject a larger share of the relevance.
Apart from this drawback one can see in Figure \ref{fig:fail} that LRP responses can be well interpreted as positive evidence (red color) and negative evidence (blue color) of a classification decision. In particular, when backpropagating the (artificial) classification decision that the image has been classified as `9', LRP provides a very intuitive explanation, namely that in the left upper part of the image the missing stroke closing the loop (blue color) speaks against the fact that this is a `9' whereas the missing stroke in the left lower part of the image (red color) supports this decision. The deconvolution method does not allow such interpretation, because it loses negative evidence while backprojecting over ReLU layers and does not use image specific information.

\section{Evaluating Heatmaps}
\label{sec:heatmap}

\subsection{What makes a good heatmap?}
Although humans are able to intuitively assess the quality of a heatmap by matching with prior knowledge and experience of what is regarded as being relevant, defining objective criteria for heatmap quality is very difficult. In this paper we refrain from mimicking the complex human heatmap evaluation process which includes attention, interest point models and perception models of saliency \cite{itti2000saliency, heeger1996computational, simoncelli2001natural, itti2001computational}, because we are interested in the relevance of the heatmap for the classifier decision. We use the classifier's output and a perturbation method to objectively assess the quality (see Section \ref{sec:framework}).
When comparing different heatmapping algorithms one should be aware that heatmap quality does not only depend on the algorithms used to compute a heatmap, but also on the performance of the classifier, whose efficiency largely depends on the model used and the amount and quality of available training data. A random classifier will provide random heatmaps.
Also, if the training data does not contain images of the digits `3', then the classifier can not know that the absence of strokes in the left part of the image (see example in Figure \ref{fig:fig1}) is important for distinguishing the digit `3' from digits `8' and `9'. Thus, explanations can only be as good as the data provided to the classifier. 

Furthermore, one should keep in mind that a heatmap always represents the classifier's view, i.e., explanations neither need to match human intuition nor focus on the object of interest. A heatmap is not a segmentation mask (see Figure \ref{fig:fig1}), on the contrary missing evidence or the context may be very important for classification. Also image statistics may be highly discriminative, i.e., evidence for a class does not need to be localized. From our experience heatmaps become more intuitive with higher classification accuracy (see Section \ref{sec:quality}), but there is no guarantee that human and classifier explanations match.
Regarding general quality criteria we believe that heatmaps should have low ``complexity'', i.e., be as sparse and non-random as possible. Only the relevant parts in the images should be highlighted and not more. We use complexity, measured in terms of image entropy or the file size of the compressed heatmap image, as an auxiliary criteria to assess heatmap quality. 

\subsection{Salient Features vs. Individual Explanations}
Salient features represent average explanations of what distinguishes one image category from another. For individual images these explanations may be meaningless or even wrong.
For instance, salient features for the class `bicycle' may be the wheels and the handlebar. However, in some images a bicycle may be partly occluded so that these parts of a bike are not visible. In these images salient features fail to explain the classifier's decision (which still may be correct). Individual explanations on the other hand do not target the ``average case'', but focus on the particular image and may identify other parts of the bike or the context (e.g., presence of cyclist) as being good explanations of the classifier's decision.

\subsection{Heatmap Evaluation Framework}
\label{sec:framework}
To evaluate the quality of a heatmap we consider a greedy iterative procedure that consists of measuring how the class encoded in the image (e.g. as measured by the function $f$) disappears when we progressively remove information from the image $\x$, a process referred to as {\it region perturbation}, at the specified locations. The method is a generalization of the  approach presented in \cite{bach15}, where the perturbation process is a state flip of the associated binary pixel values (single pixel perturbation). The method that we propose here applies more generally to {\em any} set of locations (e.g., local windows) and any local perturbation process such as local randomization or blurring.

We define a heatmap as an ordered set of locations in the image, where these locations might lie on a predefined grid.
\begin{align}
\mathcal{O} = (\r_1,\r_2,\dots,\r_L)
\end{align}
Each location $\r_p$ is for example a two-dimensional vector encoding the horizontal and vertical position on a grid of pixels. The ordering can either be chosen at hand, or be induced by a heatmapping function $h_p = \mathcal{H}(\x,f,\r_p)$, typically derived from a class discriminant $f$ (see methods in Section \ref{sec:lrp}). 
The scores $\{h_p\}$ indicate how important the given location $\r_p$ of the image is for representing the image class. The ordering induced by the heatmapping function is such that for all indices of the ordered sequence $\mathcal{O}$, the following property holds:
\begin{align}
 (i < j) \Leftrightarrow \big(\mathcal{H}(\x,f,\r_i) > \mathcal{H}(\x,f,\r_j)\big)
\end{align}
Thus, locations in the image that are most relevant for the class encoded by the classifier function $f$ will be found at the beginning of the sequence $\mathcal{O}$. Conversely, regions of the image that are mostly irrelevant will be positioned at the end of the sequence.

We consider a region perturbation process that follows the ordered sequence of locations. We call this process {\em most relevant first}, abbreviated as MoRF. The recursive formula is:
\begin{align}
\x_\text{MoRF}^{(0)} &= \x\\
\forall~1 \leq k \leq L:~ \x_\text{MoRF}^{(k)} &= g(\x_\text{MoRF}^{(k-1)},\r_{k})\nonumber
\end{align}
where the function $g$ removes information of the image $\x_\text{MoRF}^{(k-1)}$ at a specified location $\r_{k}$ (i.e., a single pixel or a local neighborhood) in the image. 
Throughout the paper we use a function $g$ which replaces all pixels in a $9\times 9$ neighborhood around $\r_{k}$ by randomly sampled (from uniform distribution) values.
When comparing different heatmaps using a fixed $g(\x, \r_k)$ our focus is typically only on the highly relevant regions (i.e., the sorting of the $h_p$ values on the non-relevant regions is not important). The quantity of interest in this case is the area over the MoRF perturbation curve (AOPC):
\begin{align}
\mathrm{AOPC} = \frac{1}{L+1} \Big\langle \sum_{k=0}^L f(\x^{(0)}_\text{MoRF}) - 
 f(\x^{(k)}_\text{MoRF}) \Big\rangle_{p(\x)}
\end{align}
where $\langle \cdot \rangle_{p(\boldsymbol{x})}$ denotes the average over all images in the data set. An ordering of regions such that the most sensitive regions are ranked first implies a steep decrease of the graph of MoRF, and thus a larger AOPC.

\section{Experimental Results}
\label{sec:results}

In this section we use the proposed heatmap evaluation procedure to compare heatmaps computed with the LRP algorithm \cite{bach15}, the deconvolution approach \cite{DBLP:conf/eccv/ZeilerF14} and the sensitivity-based method \cite{DBLP:journals/corr/SimonyanVZ13} (Section \ref{sec:sensitivity}) to a random order baseline. Exemplary heatmaps produced with these algorithms are displayed and discussed in Section \ref{sec:subjective}. At the end of this section we briefly investigate the correlation between heatmap quality and network performance.

\subsection{Setup}
We demonstrate the results on a classifier for the MIT Places data set \cite{zhou2014learning} provided by the authors of this data set and the Caffe reference model \cite{Jia13caffe} for ImageNet. We kept the classifiers unchanged. The MIT Places classifier is used for two testing data sets. Firstly, we compute the AOPC values over 5040 images from the MIT Places testing set. Secondly, we use AOPC averages over 5040 images from the SUN397 data set \cite{DBLP:conf/cvpr/XiaoHEOT10} as it was done in \cite{zhou2014object}. We ensured that the category labels of the images used were included in the MIT Places label set. Furthermore, for the ImageNet classifier we report results on the first 5040 images of the ILSVRC2012 data set. The heatmaps are computed for all methods for the predicted label, so that our perturbation analysis is a fully unsupervised method during test stage. 
Perturbation is applied to $9\times 9$ non-overlapping regions, each covering $0.157\%$ of the image. We replace all pixels in a region by randomly sampled (from uniform distribution) values. The choice of a uniform distribution as region perturbation follows one assumption: We consider a region highly relevant if replacing the information in this region in {\it arbitrary} ways reduces the prediction score of the classifier; we do not want to restrict the analysis to highly specialized information removal schemes. In order to reduce the effect of randomness we repeat the process 10 times. For each ordering we perturb the first $100$ regions, resulting in $15.7\%$ of the image being exchanged. Running the experiments for $2$ configurations of perturbations, each with $5040$ images, takes roughly $36$ hours on a workstation with $20$ ($10 \times 2$) Xeon HT-Cores. Given the above running time and the large number of configurations reported here, we considered the choice of $5040$ images as sample size a good compromise between the representativity of our result and computing time. 

\subsection{Quantitative Comparison of Heatmapping Methods}
\label{sec:sensitivity}
\begin{figure*}[t]
\centering
\includegraphics[width=1\linewidth]{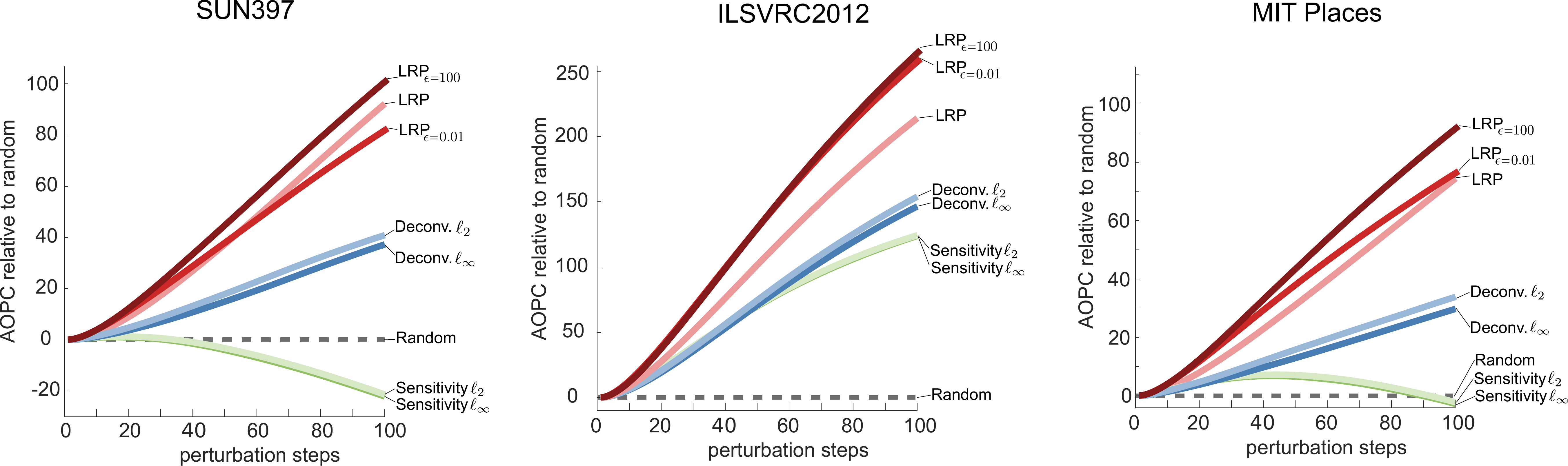}
\vskip -2mm
\caption{Comparison of the three heatmapping methods relative to the random baseline. The LRP algorithms have largest AOPC values, i.e., best explain the classifier's decision, for all three data sets.}
\label{fig:res}
\end{figure*}

We quantitatively compare the quality of heatmaps generated by the three algorithms described in Section \ref{sec:lrp}. As a baseline we also compute the AOPC curves for random heatmaps (i.e., random ordering $\mathcal{O}$). Figure \ref{fig:res} displays the AOPC values as function of the perturbation steps (i.e., $L$) relative to the random baseline. 

From the figure one can see that heatmaps computed by LRP have the largest AOPC values, i.e., they better identify the relevant (wrt the classification tasks) pixels in the image than heatmaps produced with sensitivity analysis or the deconvolution approach. This holds for all three data sets. The $\epsilon$-LRP formula (see Eq.~\ref{eq:lrp-basic}) performs slightly better than $\alpha,\beta$-LRP (see Eq.~\ref{eq:lrp-alphabeta}), however, we expect both LRP variants to have similar performance when optimizing for the parameters (here we use the same settings as in \cite{bach15}). 
The deconvolution method performs as closest competitor and significantly outperforms the random baseline.
Since LRP distinguishes between positive and negative evidence and normalizes the scores properly, it provides less noisy heatmaps than the deconvolution approach (see Section \ref{sec:subjective}) which results in better quantitative performance. As stated above sensitivity analysis targets a slightly different problem and thus provides quantitatively and qualitatively suboptimal explanations of the classifier's decision. Sensitivity provides local explanations, but may fail to capture the global features of a particular class. In this context see also the works of Szegedy \cite{DBLP:journals/corr/SzegedyZSBEGF13}, Goodfellow \cite{DBLP:journals/corr/GoodfellowSS14} and Nguyen \cite{DBLP:journals/corr/NguyenYC14} in which changing an image as a whole by a minor perturbation leads to a flip in the class labels, and in which rainbow-colored noise images are constructed with high classification accuracy.

The heatmaps computed on the ILSVRC2012 data set are qualitatively better (according to our AOPC measure) than heatmaps computed on the other two data sets.
One reason for this is that the ILSVRC2012 images contain more objects and less cluttered scenes than images from the SUN397 and MIT Places data sets, i.e., it is easier (also for humans) to capture the relevant parts of the image. Also the AOPC difference between the random baseline and the other heatmapping methods is much smaller for the latter two data sets than for ILSVRC2012, because cluttered scenes contain evidence almost everywhere in the image whereas the background is less important for object categories.

An interesting phenomenon is the performance difference of sensitivity heatmaps computed on SUN397 and MIT Places data sets, in the former case the AOPC curve of sensitivity heatmaps is even below the curve computed with random ranking of regions, whereas for the latter data set the sensitivity heatmaps are (at least initially) clearly better.
Note that in both cases the same classifier \cite{zhou2014learning}, trained on the MIT Places data, was used. The difference between these data sets is that SUN397 images lie outside the data manifold (i.e., images of MIT Places used to train the classifier), so that partial derivatives need to explain local variations of the classification function $f(x)$ in an area in image space where $f$ has not been trained properly. This effect is not so strong for the MIT Places test data, because they are much closer to the images used to train the classifier. Since both LRP and deconvolution provide global explanations, they are less affected by this off-manifold testing.

We performed above evaluation also for both Caffe networks in training phase, in which the dropout layers were active. The results are qualitatively the same to the ones shown above.
The LRP algorithm, which was explicitly designed to explain the classifier's decision, performs significantly better than the other heatmapping approaches. We would like to stress that LRP does not artificially benefit from the way we evaluate heatmaps as region perturbation is based on a assumption (good heatmaps should rank pixels according to relevance wrt to classification) which is independent of the relevance conservation principle that is used in LRP. Note that LRP was originally designed for binary classifiers in which $f(x)=0$ denotes maximal uncertainty about prediction. The classifiers used here were trained with a different multiclass objective, namely that it suffices for the correct class to have the highest score. One can expect that in such a setup the state of maximal uncertainty is given by a positive value rather than $f(x)=0$. In that sense the setup here slightly disfavours LRP. However we refrained from retraining because it was important for us, firstly, to use classifiers provided by other researchers in an unmodified manner, and, secondly, to evaluate the robustness of LRP when applied in the popular multi-class setup. 

\begin{figure*}[t]
\centering
\includegraphics[width=0.8\linewidth]{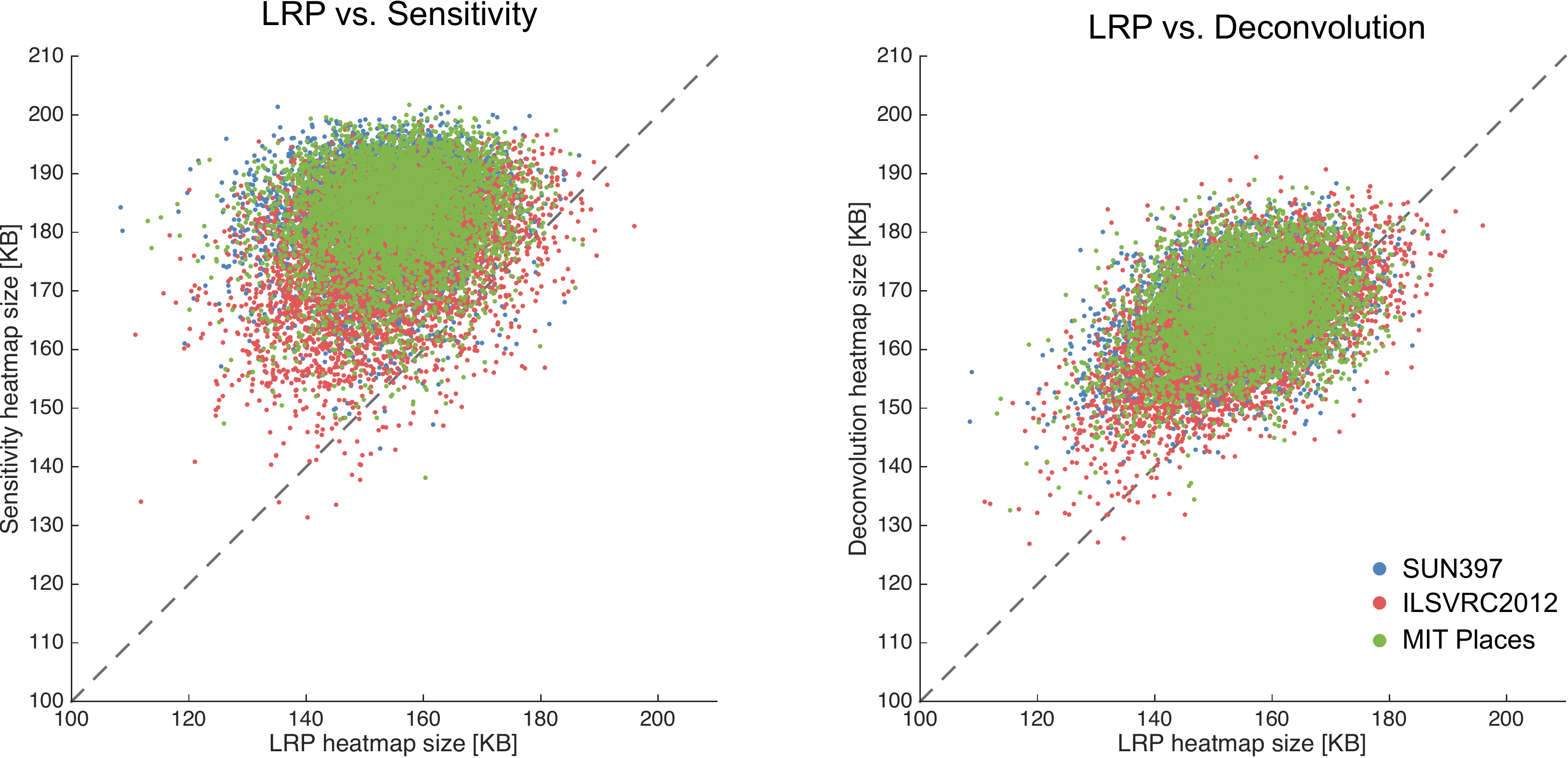}
\vskip 4mm
\includegraphics[width=0.8\linewidth]{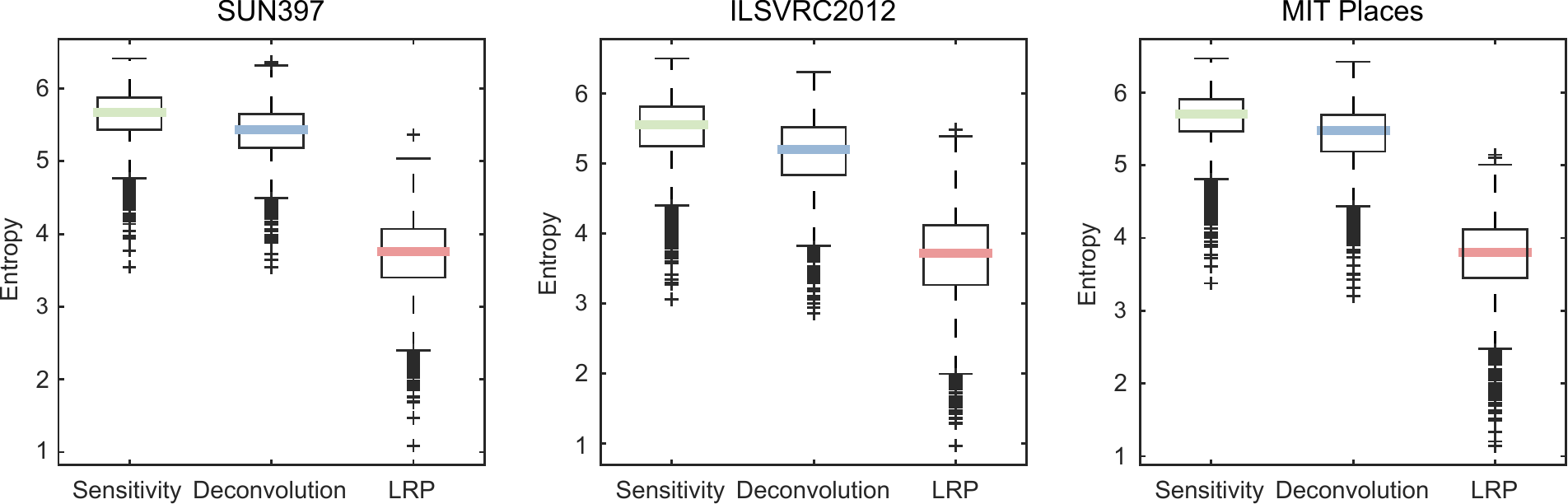}
\vskip -2mm
\caption{Comparison of heatmap complexity, measured in terms of file size (top) and image entropy (bottom). }
\label{fig:complexity}
\end{figure*}

As stated before heatmaps can also be evaluated wrt to their complexity (i.e., sparsity and randomness of the explanations). Good heatmaps highlight the relevant regions and not more, whereas suboptimal heatmaps may contain lots of irrelevant information and noise. In this sense good heatmaps should be better compressible than noisy ones. Table \ref{tab:filesize} compares the average file size of heatmaps (saved as png and jpeg (quality 90) images) computed with the three methods. The file sizes reflect the performance reported in Figure \ref{fig:res}, i.e., LRP has best performance on all three data sets and its heatmaps have smallest file size (which means that they are well compressible, i.e., have low complexity). The second best method is the deconvolution algorithm and the sensitivity approach performs worst wrt to both measures. These differences in file size are highly significant. The scatter plots in Figure \ref{fig:complexity} show that for almost all images of the three data sets the LRP heatmap png files are smaller (i.e., less complex) than the corresponding deconvolution and sensitivity files (same holds for jpeg files). Additionally, we report results obtained using another complexity measure, namely MATLAB's \texttt{entropy} function. Also according to this measure LRP heatmaps are less complex (see boxplots in Figure \ref{fig:complexity}) than heatmaps computed with the sensitivity and deconvolution methods. 

\begin{table}
\centering
\caption{Comparison of average file size of the heatmap images in KB. LRP heatmaps have smallest size, i.e., lowest complexity.}
\footnotesize
\begin{tabular}{|l|ccc|}
\hline
Method& Sensitivity $\ell_2$ & Deconvolution $\ell_2$ & LRP \\
\hline
\hline
SUN397& 183 & 166 & 154\\
ILSVRC2012& 177 & 164 & 154\\
MIT Places& 183 & 167 & 155\\
\hline
SUN397& 25 & 20 & 14\\
ILSVRC2012& 22 & 18 & 13\\
MIT Places& 25 & 20 & 14\\
\hline
\end{tabular}
\label{tab:filesize}
\end{table}

\subsection{Qualitative Comparison of Heatmapping Methods}
\label{sec:subjective}
\begin{figure*}
\centering
\includegraphics[width=0.9\linewidth]{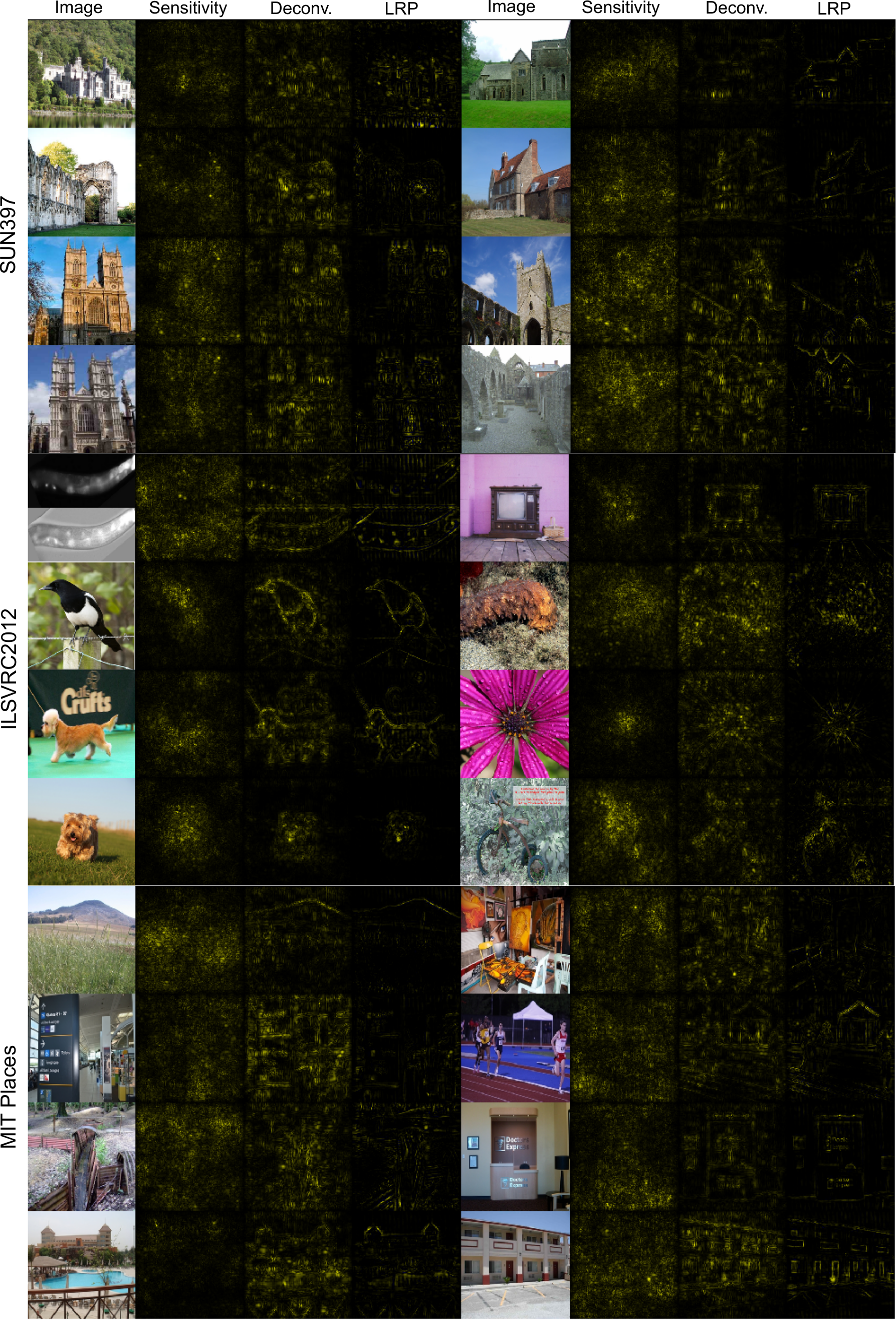}
\vskip -2mm
\caption{Qualitative comparison of the three heatmapping methods.}
\label{fig:res3}
\end{figure*}

In Figure \ref{fig:res3} the heatmaps of the first 8 images of each data set are visualized. The quantitative result presented above are in line with the subjective impressions.
The sensitivity and deconvolution heatmaps are nosier and less sparse than the heatmaps computed with the LRP algorithm, reflecting the results obtained in Section \ref{sec:sensitivity}.
For SUN 397 and MIT Places the sensitivity heatmaps are close to random, whereas both LRP and deconvolution highlight some structural elements in the scene. We remark that this bad performance of sensitivity heatmaps does not contradict results like \cite{DBLP:journals/corr/SzegedyZSBEGF13,DBLP:journals/corr/GoodfellowSS14}. In the former works, an image gets modified as a whole, while in this work we are considering the quality of selecting local regions and ordering them. Furthermore gradients require to move in a very particular direction for reducing the prediction while we are looking for most relevant regions in the sense that changing them in any kind will likely destroy the prediction.
The deconvolution and LRP algorithms capture more global (and more relevant) features than the sensitivity approach.

\subsection{Heatmap Quality and Neural Network Performance}
\label{sec:quality}
\begin{figure}[t]
\centering
\includegraphics[width=1\linewidth]{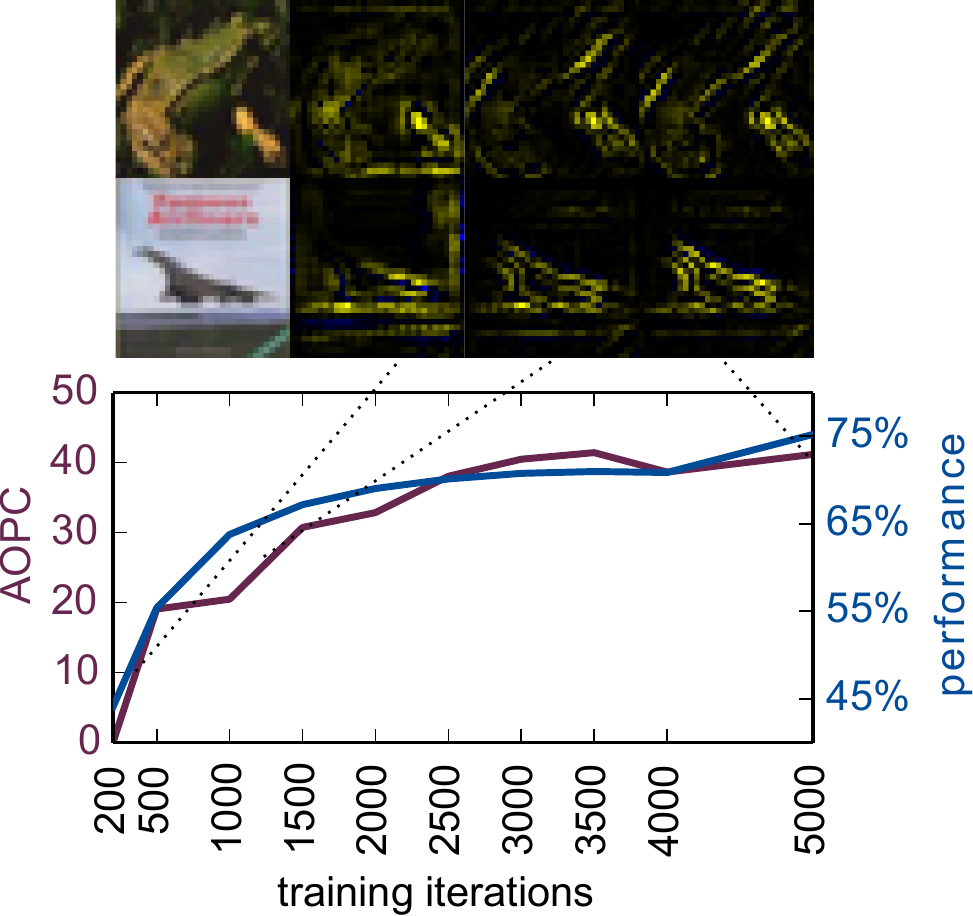}
\vskip -2mm
\caption{Evaluation of network performance by using AOPC on the CIFAR-10 data set}
\label{fig:exp4b}
\end{figure}

In the last experiment we briefly show that the quality of a heatmap, as measured by AOPC, provides information about the overall DNN performance. The intuitive explanation for this is that well-trained DNNs much better capture the relevant structures in an image, thus produce more meaningful heatmaps than poorly trained networks which rather rely on global image statistics. Thus, by evaluating the quality of a heatmap using the proposed procedure we can potentially assess the network performance, at least for classifiers that were based on the same network topology. Note that this procedure is based on perturbation of the input of the classifier with the highest predicted score. Thus this evaluation method is purely unsupervised and does not require labels of the testing images.
Figure \ref{fig:exp4b} depicts the AOPC values and the performance for different training iterations of a DNN for the CIFAR-10 data set \cite{krizhevsky2009learning}. We did not perform these experiments on a larger data set since the effect can still be observed nicely in this modest data size. The correlation between both curves indicates that heatmaps contain information which can potentially be used to judge the quality of the network. This paper did not indent to profoundly investigate the relation between network performance and heatmap quality, this is a topic for future research.

\section{Conclusion}
\label{sec:conclusion}
Research in DNN has been traditionally focusing on improving the quality, algorithmics or the speed of a neural network model. We have studied an orthogonal research direction in our manuscript, namely, we have contributed to furthering the understanding and transparency of the decision making implemented by a trained DNN: For this we have focused on the heatmap concept that, e.g. in a computer vision application, is able to attribute the contribution of individual pixels to the DNN inference result for a novel data sample. While heatmaps allow a better intuition about what has been learned by the network, we tackled the so far open problem of quantifying the quality of a heatmap. In this manner different heatmap algorithms can be compared quantitatively and their properties and limits can be related. We proposed a region perturbation strategy that is based on the idea that flipping the most salient pixels first should lead to high performance decay. A large AOPC value as a function of perturbation steps was shown to provide a good measure for a very informative heatmap. We also showed quantitatively and qualitatively that sensitivity maps and heatmaps computed with the deconvolution algorithm are much noisier than heatmaps computed with the LRP method, thus are less suitable for identifying the most important regions wrt the classification task. Above all we provided first evidence that heatmaps may be useful for assessment of neural network performance. Bringing this idea into practical application will be a topic of future research.
Concluding, we have provided the basis for an accurate quantification of heatmap quality. 

Note that a good heatmap can not only be used for better understanding of DNNs but also for a priorization of image regions. Thus, regions of an individual image with high heatmap values could be subjected to more detailed analysis. This could in the future allow highly time efficient processing of the data only {\it where it matters}. 

\bibliographystyle{IEEEtran}
\bibliography{paper}

\appendix
\section*{Choosing a perturbation method}
An ideal region perturbation method effectively removes information without introducing spurious structures.
Furthermore it neither significantly disrupts image statistics nor moves the corrupted image far away from the data manifold.
In the following we propose four different region perturbation functions $g(\x, \r_p)$:
\\[+6px]
{\bf Uniform}: replaces each pixel near $\r_p$ by a RGB value sampled from an uniform distribution $\mathcal{U}$.\\[+3px]
{\bf Dirichlet}: replaces each pixel near $\r_p$ by a RGB value sampled from a four-dimensional Dirichlet distribution $\mathcal{D}$. By sampling from this distribution we retain the image statistics.\\[+3px]
{\bf Constant}: replaces each pixel near $\r_p$ by a constant value. Here we use the average RGB value computed over all images at this location. Note that this does not mean that all pixels near $\r_p$ are replaced by the same value.  \\[+3px]
{\bf Blur}: blurs the pixels near $\r_p$ with a Gaussian filter with $\sigma = 3$. This is the only method which retains local information. \\[+6px] 

Note that using an uniform distribution for sampling from $[0,1]^3$ implies a mean of $0.5$ and a standard deviation $0.5\sqrt{1/3}$ for each pixel. However, we observed for those images which we analyzed that the image mean for certain pixels goes above $0.55$ which is not surprising, as pixels in the top of the images often show bright sky. One possibility for retaining the natural image statistics is to average RGB value computed over all images at each location (see method ``Constant'').  Another possibility is to sample from a distribution which fits the data well. In case of one pixel with one color channel, a natural choice would be to fit a beta distribution. The beta distribution generalizes for higher dimensions to a Dirichlet distribution. Since the three dimensional Dirichlet distribution does not fit the required condition $r,g,b \in [0,1], \, r+g+b \in [0,3]$, we derive a sampling scheme based on a modified four dimensional Dirichlet distribution.

\section*{Evaluating perturbation methods}
We define an alternative region perturbation process where locations are considered in reverse order. We call this process {\em least relevant first}, or abbreviated as LeRF. In that case, the perturbation process is defined by a new recursion formula:
\begin{align}
\x_\text{LeRF}^{(0)} &= \x\\
\forall~1 \leq k \leq L:~ \x_\text{LeRF}^{(k)} &= g(\x_\text{LeRF}^{(k-1)},\r_{L+1-k}).\nonumber
\end{align}
We expect in this second case, that the class information in the image should be very stable and close to the original value for small $k$, and only drop quickly to zero as $k$ approaches $L$. 

We would like the region perturbation process to destroy the class information for highly relevant regions, and leave the class intact for least relevant regions. The idea behind it is to maintain a balance between staying on the data manifold and being able to alter the image in a way which can be sensed by the classifier. To quantify this property, we propose to monitor the gap between the two region perturbation processes LeRF and MoRF:
\begin{align}
\mathrm{ABPC} = \frac{1}{L+1} \Big\langle \sum_{k=0}^L f(\x^{(k)}_\text{LeRF}) - f(\x^{(k)}_\text{MoRF}) \Big\rangle_{p(\x)}
\end{align}
where $f$ is a class scoring function (not necessarily the same as the one used for computing the heatmaps), and where $\langle \cdot \rangle_{p(\x)}$ measures the expectation of the measured quantity with respect to the distribution of input images. The area between perturbation curves (ABPC) is an indicator of how good a heatmapping technique is, how good the class scoring function $f$ that induces the heatmaps is, and also how good a region perturbation process is at removing local information in the image without introducing spurious structure in it. Large ABPC values are desirable.

\section*{Comparison of perturbation methods on SUN397}
\begin{figure}[t]
\centering
\includegraphics[width=0.9\linewidth]{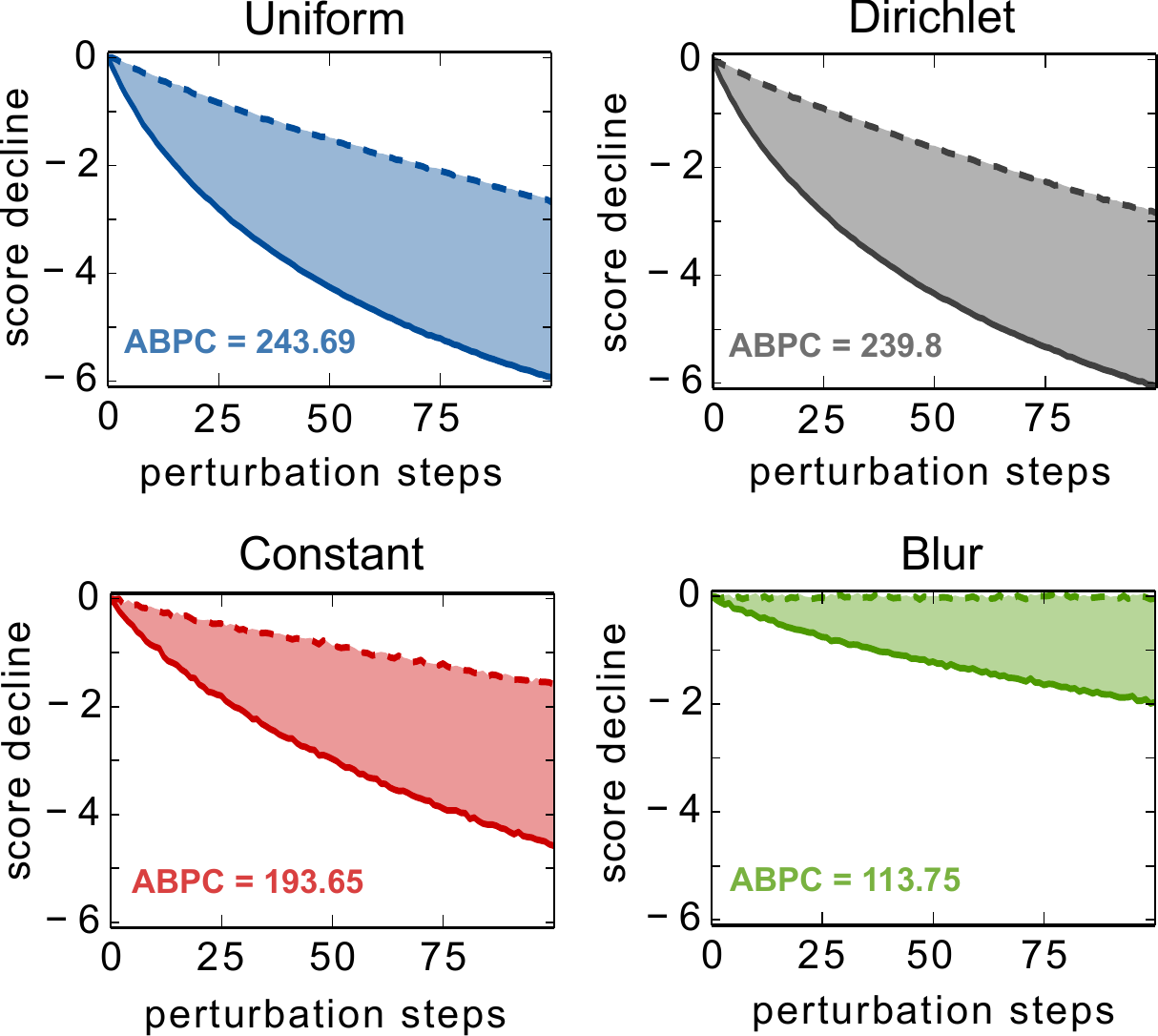}
\vskip -2mm
\caption{Comparison of the four image corruption schemes on the SUN397 data set. Dashed and solid lines are LeRF and MoRF curves, respectively.}
\label{fig:fig6}
\end{figure}

Figure \ref{fig:fig6} depicts the results of this comparison for the four functions $g$ presented before when applied to the SUN397 data set. For each test image we compute heatmaps using LRP. Subsequently, we measure the classification output from the highest linear layer while continuously removing the most and least relevant information, respectively. One can see in Figure \ref{fig:fig6} that blurring fails to remove information. The MoRF curve is relatively flat, thus the DNN does not lose the ability to classify an image even though information from relevant regions is being continuously corrupted by $g(\x, \r_k)$. Similarly, replacing the pixel values in $\r_k$ by constant values does not abruptly decrease the score. In both cases the DNN can cope with losing an increasing portion of relevant information. The two random methods, Uniform and Dirichlet, effectively remove information and have significantly larger ABPC values.	
Although the curves of {\it both} region perturbation processes, MoPF and LePF, show a steeper decrease than in case of Constant and Blur, the relative score decline is much larger resulting in larger ABPC values.

\end{document}